\documentclass[letterpaper]{article}

\usepackage{natbib,alifeconf}  
\usepackage{amsmath, amssymb}
\usepackage{url,hyperref,cleveref}
\usepackage{booktabs}
\usepackage{hhline} 
\usepackage{colortbl} 
\usepackage{xcolor} 

\renewcommand{\S}[0]{\mathcal{S}}
\newcommand{\M}[0]{\mathcal{M}}
\newcommand{\B}[0]{\mathcal{B}}
\newcommand{\A}[0]{\mathcal{A}}

\newcommand{\BB}[0]{\mathbb{B}}
\renewcommand{\AA}[0]{\mathbb{A}}

\usepackage{tikz}
\usetikzlibrary{shapes.geometric, positioning, arrows.meta, bending,backgrounds,calc}
\usetikzlibrary{decorations.markings}

\title{ALIFE2024}

\title{Modeling language contact with the Iterated Learning Model}

\author{
    Seth Bullock \and
    Conor Houghton,
    \mbox{}\\
    Intelligent Systems Laboratory, University of Bristol, United Kingdom
    conor.houghton@bristol.ac.uk
} 

\begin{document}

\maketitle

\begin{abstract}

Contact between languages has the potential to transmit vocabulary and other language features; however, this does not always happen. Here, an iterated learning model is used to examine, in a simple way, the resistance of languages to change during language contact. Iterated learning models are agent-based models of language change, they demonstrate that languages that are expressive and compositional arise spontaneously as a consequence of a language transmission bottleneck. A recently introduced type of iterated learning model, the Semi-Supervised ILM is used to simulate language contact. These simulations do not include many of the complex factors involved in language contact and do not model a population of speakers; nonetheless the model demonstrates that the dynamics which lead languages in the model to spontaneously become expressive and compositional, also cause a language to maintain its core traits even after mixing with another language.
\end{abstract}

\section{Introduction}

Iterated Learning Models (ILMs) are agent-based models in which languages evolve as they are  transmitted from generation to generation. They were proposed as a description of how languages become compositional and expressive \citep{Kirby2001,KirbyHurford2002}. Here, the Semi-Supervised ILM, a specific version of the model recently introduced in \cite{BunyanBullockHoughton2024}, is used to simulate the effect on a language of contact with another language.

The continuous processes required to produce the vast array of languages currently in use seems to indicate that languages are extremely mutable and unstable. Languages certainly change. For example, since Jonathan Swift's novel \textsl{Gulliver's Travels} was published less than three hundred years ago on 28 October 1726, the English lexicon has expanded enormously to provide a vocabulary for the many novel objects, behaviours and social structures that have resulted from the Industrial Revolution, the Age of Revolution and the Computer Age. However, for all that, the language has not changed that much: though unarguably `old-fashioned' in its phrasing, \textsl{Gulliver's Travels} is perfectly comprehensible to the contemporary reader. 

This is striking. Between Swift's era and the present day, English has become a \textsl{lingua franca}, spoken for periods as the language of administration in countries within which it was not native and used as a second language by more people than use it as a first language. Consequently, it might be expected that it would have changed dramatically with a huge number of words from other languages crowding out older words and rendering the novel incomprehensive. Indeed the idea that English continually adds new words by taking them from other languages is something of a cliche. 

However, it is not really true. Taking the example of English and Irish (endonym: Gaeilge, ISO 639-3: gle); Jonathon Swift was among the first prominent Irish writers writing in English and there has been a rich history of interaction between speakers of Irish and of English in the period since \textsl{Gulliver's Travels} was written. Ultimately an English language has replaced Irish as the main language of Ireland and the descendants of the substantial Irish diaspora are for the most part English speakers. Nonetheless only a handful of words derived from Irish are in common use in English.  Indeed, while it is claimed that the aboriginal Celtic languages of Great Britain and Ireland form a language substrate for English \citep{FilppulaKlemplaPaulasto2008}, even in Ireland  any more recent influence of Irish on the grammar, pronunciation and lexicon of the language now spoken is apparent only in small, though valued, ways. In fact, as in the example of the word `oxter' for `armpit', many of the lexical difference between Irish-English and other Englishes represent the preservation in Irish-English of English words that have fallen out of use elsewhere, rather than the influence of Irish.

Of course, the dynamics of language contact is complex and it is certainly not invariably true that one language retreats while leaving the other unchanged. Indeed, in the case of English, the Norman Conquest and the influence of Old French (ISO 639-3: fro) caused substantial changes, marked by the transition from Old English (ISO 693-3: ang) to Middle English (ISO 693-3: enm). Old French in turn showed considerable influences from Old Frankish (ISO 639-3: frk), though this is a salutary example; the main effect of Old Frankish was to change stress pattern in pronunciation, disrupting the grammatical information in word endings that Vulgar Latin relied on, leading to a shift from a synthetic to an analytic language. In other words, although contact with Old Frankish was important in the change from Vulgar Latin to Old French, these changes did not mean Old French resembled Old Frankish any more than Vulgar Latin did \cite{Rickard2003,Lodge2013}. 

The various outcomes of language contact depend on multiple factors such as prestige, economic opportunity and the need for new words for phenomena and objects familiar to speakers of one language but not the other \citep{ThomasonKaufman1988,Mufwene2001,Matras2007,SakelMatras2008}. Our purpose here is to examine the, perhaps surprising, stability of languages during asymmetric contact from the simplest possible point-of-view. We explore whether the mechanism underpinning the development of compositionality and expressivity in the Semi-Supervised ILM could also account for the stability of one of the two languages in a contact situation.

\subsection{The Iterated Language Model}

Languages are made easier to learn, and easier to use, by their properties of \textsl{compositionality} and \textsl{expressivity} \citep{Oliphant1999}. A compositional language is one in which each element of meaning is encoded consistently by the same morpheme within the signal, irrespective of context, so `violin' means much the same thing in `she plays the violin' and `it is violin shaped'. While all human languages exhibit compositionality of some kind, they are not completely compositional, so, in other contexts the word `fiddle' can be used: `Mair\'{e}ad N\'{i} Mhaonaigh is a leading exponent in the Donegal fiddle tradition'. An expressive language is one in which a range of different meanings are unambiguously encoded by different signals, so the viola is called a `viola' in English and can be distinguished by name from a `violin'. However, just as real languages are not perfectly compositional, they are not perfectly expressive: the `fiddle' in `it is tempting to fiddle with it, even though it is perfect as it is' does not refer to a violin. Along with expressivity and compositionality, a third simple property required from a language is \textsl{stability}; while languages change, to support communication they should not change too much from one generation to the next.

ILMs are a family of computational models designed to simulate language change. They were originally intended to describe a mechanism by which languages can become expressive, compositional and stable over multiple generations of use \citep{Kirby2001,KirbyHurford2002}. In an ILM there is a language-using `tutor' agent and a language-learning `pupil' agent. The adult teaches language to the pupil by exposing it to a set of utterances. The pupil then becomes a tutor and teaches a new pupil in the same fashion. A language learning bottleneck is a key feature of these models: since each tutor only presents a finite subset of the language to their pupil, when the pupil in turn becomes an adult it will need to generate utterances that it has not experienced before. The process of generalization inherent in the model is thought to be critical to the behaviour of the ILM.

In the version of the ILM considered here a language consists of a set of meaning, $\M$, and a set of signals, $\S$, both represented by binary vectors, typically here these will be length ten vectors. In addition the agent has an encoder map $e:\M\rightarrow\S$ and a decoder map $d:\S\rightarrow\M$ mapping signals to meanings and vice versa, respectively:
\begin{center}
   \begin{tikzpicture}
  \node (S) at (0,0) {$\S$};
  \node (M) at (4,0) {$\M$};
  \draw[-{Latex[bend]}, bend right] (M) to node[above] {$e$} (S);
  \draw[-{Latex[bend]}, bend right] (S) to node[below] {$d$} (M);
\end{tikzpicture}
\end{center}
A language, in this context, corresponds to a map from meanings to signals and so an agent's language can be thought of as being equivalent to its encoder. In \cite{KirbyHurford2002} the decoder $d$ is a simple neural network whereas the encoder $e$ is a look-up table listing all possible meanings and their corresponding signal. During learning a `bottleneck' subset of meanings $\B\in\M$ is chosen and the tutor presents this training meaning-signal pairs $\{(e(b),b)|b\in\B\}$. This is used to train the pupil's decoder. At the end of training, when the pupil `grows up', an inversion process, known as \textsl{obversion} \citep{OliphantBatali1997}, is used to build an encoder for this agent, after which it can become a tutor for a newly created na\"ive pupil. 

During obversion, the pupil's decoder is used to assign a probability to every possible signal-meaning pair, the encoder map from meaning to signal is then calculated by a process of `picking winners'. This is computationally expensive to the extent that it makes simulations impractical for languages with bit length greater than about ten. It is also unrealistic as it imagines that a pupil is incapable of an utterances until it matures, at which point it must consider all possible signals and all possible meaning and match them to calculate the encoder that it will rely on in order to be able to communicate as an adult tutor.

\begin{figure}[tb]
    \centering
    \begin{tikzpicture}[>=Stealth, node distance=0.2cm, every node/.style={circle,draw,minimum size=9.5mm}]

\tikzstyle{invisibleNode} = [inner sep=0, outer sep=0, minimum size=1mm, draw=none]

\tikzset{
  halfway arrow/.style={
    decoration={markings, mark=at position 0.5 with {\arrow{>}}},
    postaction={decorate}
  }
}

\node [fill=cyan](I-1) {$m_1$};
\node [fill=cyan](I-2) [below=of I-1] {$m_2$};
\begin{scope}[on background layer]
\node [draw=none, below=0.2cm of I-2] (I-3) {$\vdots$}; 
\end{scope}
\node [fill=cyan](I-4) [below=0.2cm of I-3] {$m_n$};
\node [invisibleNode] (I-5) [above=0.5cm of I-1] {};
\node [invisibleNode] (I-6) [above=0.5cm of I-5] {};

\node [fill=cyan](H1-1) [right=of I-1, xshift=0.2cm] {};
\node [fill=cyan](H1-2) [below=of H1-1] {};
\begin{scope}[on background layer]
\node [draw=none, below=0.2cm of H1-2] (H1-3) {$\vdots$}; 
\end{scope}
\node [fill=cyan](H1-4) [below=0.2cm of H1-3] {}; 
\node [invisibleNode] (H1-5) [above=0.5cm of H1-1] {};
\node [invisibleNode] (H1-6) [above=0.3cm of H1-5] {};

\node [fill=green](M-1) [right=of H1-1, xshift=0.2cm] {$s_1$};
\node [fill=green](M-2) [below=of M-1] {$s_2$};
\begin{scope}[on background layer]
\node [draw=none, below=0.2cm of M-2] (M-3) {$\vdots$};
\end{scope}
\node [fill=green](M-4) [below=0.2cm of M-3] {$s_n$};
\node [invisibleNode] (M-5) [above=0.5cm of M-1] {};
\node [invisibleNode] (M-6) [above=0.3cm of M-5] {};

\node [fill=yellow](H2-1) [right=of M-1, xshift=0.2cm] {};
\node [fill=yellow](H2-2) [below=of H2-1] {};
\begin{scope}[on background layer]
\node [draw=none, below=0.2cm of H2-2] (H2-3) {$\vdots$}; 
\end{scope}
\node [fill=yellow](H2-4) [below=0.2cm of H2-3] {}; 
\node [invisibleNode] (H2-5) [above=0.5cm of H2-1] {};
\node [invisibleNode] (H2-6) [above=0.3cm of H2-5] {};

\node [fill=yellow](O-1) [right=of H2-1, xshift=0.2cm] {$m'_1$};
\node [fill=yellow](O-2) [below=of O-1] {$m'_2$};
\begin{scope}[on background layer]
\node [draw=none, below=0.2cm of O-2] (O-3) {$\vdots$}; 
\end{scope}
\node [fill=yellow](O-4) [below=0.2cm of O-3] {$m'_n$};
\node [invisibleNode] (O-5) [above=0.5cm of O-1] {};
\node [invisibleNode] (O-6) [above=0.5cm of O-5] {};

\begin{scope}[on background layer]
\foreach \i in {1,2,4} {
    \foreach \j in {1,2,4} {
        \draw[->] (I-\i) -- (H1-\j);
        \draw[->] (H1-\i) -- (M-\j);
        \draw[->] (M-\i) -- (H2-\j);
        \draw[->] (H2-\i) -- (O-\j);
    }
}

\node (SI)[align=center, text width=2.5cm, below=of I-3, draw=none] {meaning};
\node (M)[align=center, text width=2.5cm, below=of M-3, draw=none] {signal};
\node (SO)[align=center, text width=2.5cm, below=of O-3, draw=none] {meaning};

\end{scope}

\draw[halfway arrow] (I-6) -- node[invisibleNode, midway, above=2mm, sloped] {$a$} (O-6);
\draw[halfway arrow] (I-5) -- node[invisibleNode, midway, above=2mm, sloped] {$d$} (M-5);
\draw[halfway arrow] (M-5) -- node[invisibleNode, midway, above=2mm, sloped] {$e$} (O-5);


\end{tikzpicture}
    \caption{\textbf{Neural network.} In the Semi-Supervised ILM both the encoder, $e$ (yellow), and decoder, $d$ (blue), are neural networks; in $e$ the meaning vector forms the input, it is mapped to hidden layer and then to a signal vector; in $d$ a signal vector forms the input, there is again a hidden layer and the output is a meaning vector. The autoencoder, $a$, is formed by concatenating the two neural networks, with the signal layer, green, in this case forming a third hidden layer for the autoencoder.}
    \label{fig:nn}
\end{figure}
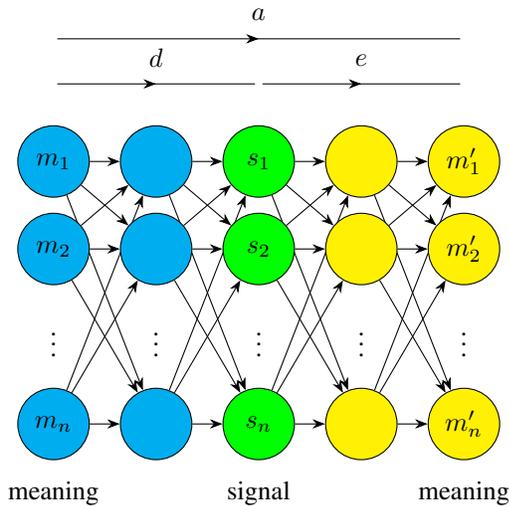

In a recent paper, \cite{BunyanBullockHoughton2024} introduce a new ILM which is capable of demonstrating the emergence of expressivity, compositionality and stability under bottlenecked conditions without the need for obversion. In this ILM, both the decoder and the encoder are neural networks. During learning the tutor presents pairs $\{(e(b),b)|b\in\B\}$ to both the pupil's decoder and to its encoder. In addition, the pupil creates a third neural network, $a=d\circ e$, by concatenating the encoder and decoder. This third network is presented with meanings from a set $\A\in\M$ and it learns as an autoencoder; this is intended to mimic a process of internal learning in which a child sees objects and events and imagines how they might be expressed in its language. This is illustrated in Fig.~\ref{fig:nn}. In \cite{BunyanBullockHoughton2024} it is noted that this ILM is considerably less computationally burdensome and it is argued that it is more realistic as a model of language development. This new version of the ILM is used here and is referred to as the Semi-Supervised ILM.

\subsection{Modeling language contact}

The ILM was proposed to model the way in which a language can acquire expressivity, compositionality and stability through iterated learning with a bottleneck. Here though, another case is considered in which there are two pre-existing languages; the $\AA$ language and the $\BB$ language, each of which are both expressive and compositional. In our iterated language model the first pupil will be taught a mixture of these two languages, so after the training set of meanings $\B$ has been selected from $\M$ some meanings will be encoded according to the $\AA$ language and some according to the $\BB$ language: specifically for each meaning in $\B$ the $\AA$ language signal will be selected with probability $p$ and, conversely, the $\BB$ language meaning with probability $1-p$. This is intended to reflect a difference in the number of exemplars presented to a child in a language contact situation, or, more generally, $p$ is supposed to represent any weighting, such as the fraction of $\AA$ examplars or the authority and emphasis with which those examples are presented. After this first initial generation the language is evolved through the usual iterated learning process and its similarity, $a$ and $b$, to the two original languages is quantified. The calculation of these similarities is discussed below.

\section{Methods}

\begin{table}
\begin{center}
 \arrayrulecolor{black}
\begin{tabular}{|l|l|}
\arrayrulecolor{black}
\hhline{|-|-|}
     $n_1$&bit length of the meaning vectors.\\
     $n_2$&number of nodes in the hidden layers.\\
     $n_3$&bit length of the signal vectors.\\
\arrayrulecolor{gray}
\hhline{|-|-|}     
     $n_1\times n_2\times n_3$&agent with a $n_1\times n_2\times n_3$ encoder\\
     &and a $n_3\times n_2\times n_1$ decoder.\\
\arrayrulecolor{gray}
\hhline{|-|-|}
     $\M$ & set of meanings.\\
     $\S$ & set of signals.\\
     $\B \subset \M$ &used for supervised learning.\\
     $\A \subset \M$ & used for unsupervised learning.\\
     \arrayrulecolor{gray}
\hhline{|-|-|}
$\AA$ and $\BB$&the two initial languages.\\
$p$&the language mixing parameter.\\
     \arrayrulecolor{gray}
\hhline{|-|-|}
     $d:\S\rightarrow\M$&decoder map.\\
     $e:\M\rightarrow\S$&encoder map.\\
     $a:\M\rightarrow\M$&autoencoder map $a=d\circ e$.\\
\arrayrulecolor{gray}
\hhline{|-|-|}
     $x$&expressivity (blue).\\
     $c$&compositionality (orange).\\
     $s$&stability (maroon in graphs).\\
    $a$&similarity to language $\AA$ (red).\\
    $b$&similarity language $\BB$ (red).\\
\arrayrulecolor{black}
\hhline{|-|-|}     
\end{tabular} 
\end{center}
\caption{\textbf{Glossary}. Here a \textsl{meaning} is a $n_1$ bit length binary vector whose components are facts or elements of meaning and a \textsl{signal} is a $n_3$ bit length binary vector whose components are morphemes. There is a notation clash, $a$ is used both for the encoder map and the similarity to language $\AA$, hopefully which is which is clear in context. Another near clash is the use of $\AA$ and $\BB$ for languages and $\A$ and $\B$ for example sets, here typeface is used to make the distinction.\label{tab:glossary}}
\end{table}

The maps $d$ and $e$ are all-to-all feedforward neural networks with one hidden layer. It is convenient to introduce a short hand for the size of each layer of the encoder $e$: $n_1\times n_2\times n_3$; $n_1$ is the bit length of the meaning vectors, $n_2$ the number of nodes in the hidden layer and $n_3$ the bit length of the signal vectors, the decoder $d$, in turn, is a $n_3\times n_2\times n_1$ neural network. This, and other pieces of notation are summarized in Table~\ref{tab:glossary}.

Both the hidden layer and the output layer have sigmoid non-linearities. Since $d$ and $e$ map to binary vectors the output values are binarized by setting them to zero or one using a threshold of 0.5. However, the signal layer forms the middle layer of the autoencoder $a$ and so the rounding step is not performed on this layer in $a$: back-propagation requires that the values in the hidden layer remain continuous and, in any case, the autocoder is intended to model the child contemplating the range of different possible utterances likely to correspond to a meaning.

All weights are initialized using the default \texttt{flux.jl} Xavier initialization. The three networks are trained at the same time using stochastic gradient descent with learning rate $\eta=5.0$. The number of epochs is $20$. For each generation a bottleneck set, $\B$, and an autoencoder set  $\A$ are chosen randomly from $\M$. At the start of each epoch two copies of $\B$ are made, each in a different random order; at each iteration an item from the first copy is presented to $d$ and an item from the other to $e$; for the $10\times 10\times 10$, $10\times 12\times 10$, $9\times 11\times 12$ and  $10\times 15\times 20$ agents $r=15$ randomly chosen meanings from $\A$ are then presented to $a$, for the $20\times 30\times 20$ network $r$ is increased to 30.

Two initial languages $\AA$ and $\BB$ are formed by randomly shuffling the `identity language', the language in which each meaning vector is mapped to the identical signal vector. The shuffling is done in a way that preserves compositionality, this means that each bit in the meaning determines a different specific bit in the signal, though, unlike in the identity language, it will not, in general, map to a bit in the same position and `1' may map to a `0' rather than to `1'. In most simulations the meaning and signal space are the same length. In the example where the signal space is larger the unused bits, which after shuffling are located in specific random positions along the signal vector, are randomly set to zero or one for all signals, that is, they remain the same irrespective of meaning. The mixed language used to teach the first pupil is calculated as a look up table where for each meaning in $\M$ the encoded signal from the $\AA$ language is chosen with probability $p$, otherwise the $\BB$ language is used.

\subsection{Quantifying language properties}

A Hamming similarity is used to quantify the similarity between languages. The fraction of meanings on which they agree is calculated: 
\begin{equation}
    \tilde{f}=\frac{|\{m:e_1(m)=e_2(m)\text{ for }m\in\M\}|}{|\M|}
\end{equation}
where $e_1$ and $e_2$ are the two encoder maps. A background value, $f_0$, is calculated for random pairs of languages, the similarity is then 
\begin{equation}
f=\frac{\tilde{f}-f_0}{1-f_0},
\end{equation}
Here $a$ is used to denote the similarity from the $\AA$ language and $b$ from the $\BB$ language. In addition, expressivity, compositionality and stability are calculated; of these, stability is the most straightforward, it is the similarity between a language and the language for the previous generation. Expressivity is based on the fraction of the signal space that is employed by the language
\begin{equation}
    \tilde{x}=\frac{|\{e(m)\text{ for }m\in\M\}|}{|\S|}
\end{equation}
where $x$ is calculated from $\tilde{x}$ by removing the background value as before. It is more difficult to define a good measure of compositionality; roughly speaking the measure used here is based on taking each bit in the meaning vector and calculating the conditional entropy, over all values of the other bits in the meaning vector, of the best predicted bit in the signal. This is described in more detail in \cite{BunyanBullockHoughton2024}, in summary though, $c$, is one for a perfectly compositional language and zero if the language is no more compositional than chance.

\subsection{Simulations}

For all the simulations where the meaning vector has bit length ten $|\B|=80$ and $|\A|=240$; for the example where the meaning vector has bit length 20 a larger bottleneck is used: $|\B|=185$ and $|\A|=555$. Results for 50 simulations are presented for all examples.

\subsection{Software and libraries}
All code was written in \texttt{Julia} and run on \texttt{Julia} 1.9.3, neural networks were trained using \texttt{Flux v0.14.6} and plotting used \texttt{Gadfly} v1.3.4, \texttt{Compose} v0.9.5 and \texttt{Colors} v0.12.10. Data was handled using \texttt{DataFrames} v1.5.0 and \texttt{CSV} v0.10.11. The code used to produce the figures and the corresponding simulated data is available at \texttt{github.com/IteratedLM/2024$\_$04$\_$mix}.

\section{Results}

\begin{figure*}
    \centering
    \begin{tabular}{p{0.17\textwidth} p{0.17\textwidth} p{0.17\textwidth} p{0.17\textwidth} p{0.17\textwidth}}
        \multicolumn{5}{c}{$p=0.5$: $\AA$ and $\BB$ have equal influence}\\[10pt]
        $x$ & $c$ & $s$ & $a$ & $b$ \\
        \includegraphics[width=0.17\textwidth]{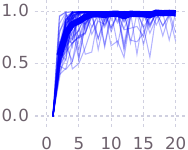} &
        \includegraphics[width=0.17\textwidth]{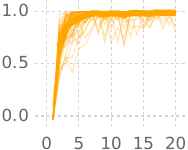} &
        \includegraphics[width=0.17\textwidth]{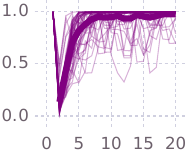} &
        \includegraphics[width=0.17\textwidth]{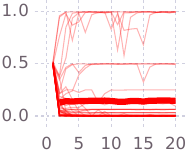} &
        \includegraphics[width=0.17\textwidth]{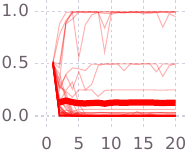} \\[15pt]
        \multicolumn{5}{c}{$p=0.55$: $\AA$ has slightly more influence than $\BB$}\\[10pt]
        $x$ & $c$ & $s$ & $a$ & $b$ \\
        \includegraphics[width=0.17\textwidth]{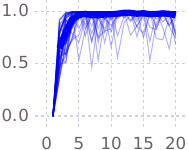} &
        \includegraphics[width=0.17\textwidth]{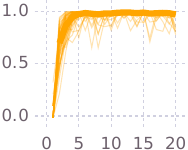} &
        \includegraphics[width=0.17\textwidth]{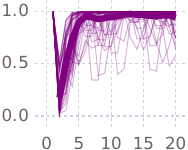} &
        \includegraphics[width=0.17\textwidth]{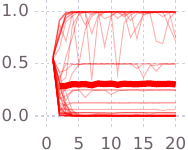} &
        \includegraphics[width=0.17\textwidth]{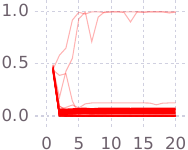} \\[15pt]
        \multicolumn{5}{c}{$p=0.75$: $\AA$ has much more influence than $\BB$}\\[10pt]
        $x$ & $c$ & $s$ & $a$ & $b$ \\
        \includegraphics[width=0.17\textwidth]{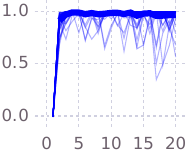} &
        \includegraphics[width=0.17\textwidth]{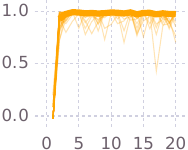} &
        \includegraphics[width=0.17\textwidth]{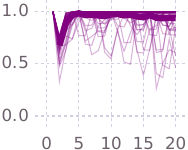} &
        \includegraphics[width=0.17\textwidth]{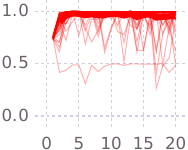} &
        \includegraphics[width=0.17\textwidth]{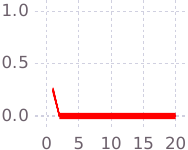} \\
        \qquad \quad generations&\qquad \quad generations&\qquad \quad generations&\qquad \quad generations&\qquad \quad generations
    \end{tabular}
    
    \caption{\textbf{Evolution of the mixed language}. Here the language is evolved for 20 generations and the results of 50 simulations are plotted for five different language attributes. In every graph the individual simulations are plotted with thin pale lines, the mean with a thicker, darker line. The first three columns plot attributes of the language itself, showing its expressivity ($x$), compositionality ($c$) and the stability ($s$) of the iterated learning. The final two columns show how similar the languages are to the initial $\AA$ and $\BB$ languages, these similarities are $a$ and $b$. The three rows present results for three different values of $p$, the fraction of the initial language taken from language $\AA$, these are $p=0.5$ when the new language is almost always different from either $\AA$ or $\BB$, $p=0.55$ when the new language is typically similar to $\AA$, sometimes to $\BB$ and sometimes to neither, and $p=0.75$ when the new language almost always resembles $\AA$. In each case a $10\times 10\times 10$ agent is used.}
    \label{fig:10_10}
\end{figure*}

Figure~\ref{fig:10_10} shows the result of language mixing for three different values of the mixing parameter $p$. The initial languages $\AA$ and $\BB$ are perfectly expressive and compositional. However, the mixed language is not and there is initially a low value for $x$ and $c$. Nonetheless the language quickly recovers these properties through iterated learning: within ten generations the languages are expressive and compositional and become stable from generation to generation. This is true for all values of $p$, however, if $p=0.5$ the language rarely ends up matching either $\AA$ or $\BB$ while for $p=0.75$ it almost invariably matches $\AA$.

\begin{figure*}
    \centering
    \begin{tabular}{ll}
    \textbf{A}&\textbf{B}\\
    \includegraphics[]{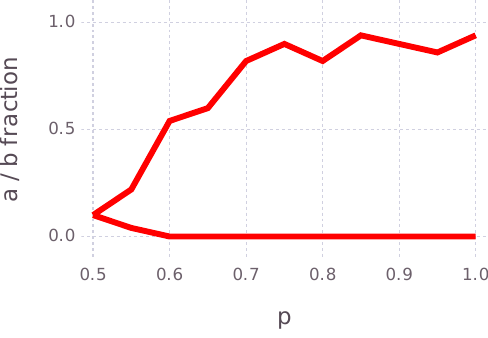}&
    \includegraphics[]{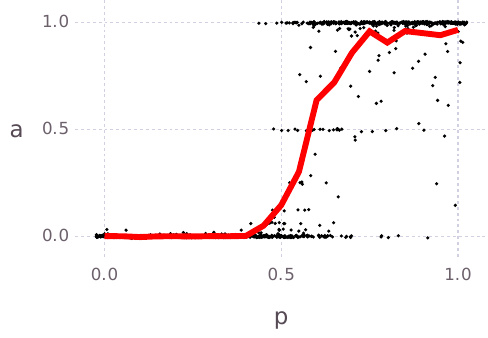}
    \end{tabular}
    \caption{\textbf{Modeling language contact}. Language contact is simulated 50 times for $p$ values from 0.5 to 1.0 in increments of 0.05. \textbf{A} plots the fraction of simulations which result in a value of $a$, upper line, or $b$, lower line, greater than 0.9. In \textbf{B} $a$ is plotted vertically, although the simulations were only run for $p\in [0.5,1.0]$ there is no distinction between the $\AA$ language and the $\BB$ language, this means that the $b$ values in $p\in [0.5,1.0]$ can be used to extend the range over which $a$ is plotted to all of $p\in[0.0,1.0]$. Each black dot corresponds to one of the 50 simulations at each $p$ value. Horizontal jitter of $\pm 0.025$ has been added to aid visualization. The red line is the average. Both \textbf{A} and \textbf{B} describe the same set of simulations using a $10\times 10\times 10$ agent.
    }
    \label{fig:p}
\end{figure*}

This is further illustrated in Fig.~\ref{fig:p} where the fraction of languages that have $a>0.9$ or $b>0.9$ is plotted against the value of $p$ (see Fig.~\ref{fig:p}\textbf{A}) and the average value of $a$ and $b$ is plotted, again, against $p$ (see Fig.~\ref{fig:p}\textbf{B}). It is apparent that there is a very rapid transition from the behaviour near $p=0.5$ where a new language is often formed, to the behaviour away from $p=0.5$ where one of the two original languages is quickly reconstructed from the elements that are present in the original mixture. What has not been assessed is the nature of the mixed languages at $p=0.5$; although typically they are not similar to either $\AA$ nor $\BB$ it is still possible that they contains some evidence of each as is the case with mixed languages like Maltese (ISO 639-3: mlt). However, Fig.~\ref{fig:p}\textbf{B} indicates that this is rare, the low average value of $a$ is not the results of a large number of trials in which the eventual value of $a$ is low, it is the result of a large number of trials giving an $a$ value of zero and a smaller number where $a=1.0$ or $a=0.5$. It will be interesting in future work to consider in more detail the language that results whem $p$ is near 0.5 and to consider more complicate multi-language mixing as happens in the formation of Creoles, though, in the case of many Creoles there are other important and challenging sociological factors involved in their formation.

\begin{figure}
    \centering
    \includegraphics[]{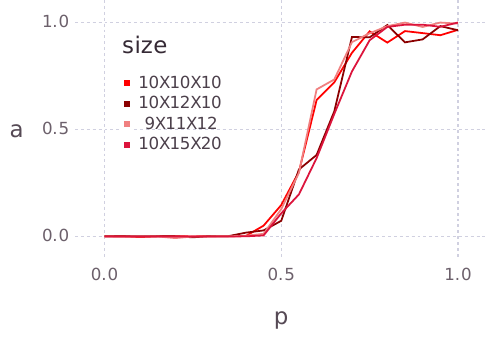}
    \caption{\textbf{The behaviour does not change substantially when the size of hidden layer or signal space is changed.} Four different architectures are compared here. Here the average value of $a$ is given for the $10\times 10\times 10$, $10\times 12\times 10$ and $9\times 11\times 12$ and $10\times 15\times 20$ agents. The four curves are similar.}
    \label{fig:compare}
\end{figure}

In Fig.~\ref{fig:compare} the performance of the $10\times 10\times 10$ Semi-Supervised ILM is compared to three other Semi-Supervised ILMs. The Semi-Supervised ILM appears to perform more robustly if the hidden layer has more nodes than the bit length of the signal and meaning spaces \cite{BunyanBullockHoughton2024}. Here, the same set of simulations as above was performed using a Semi-Supervised ILM with a $10\times 12\times 10$ agent. In addition, it might be suspected that the model returns to the $\AA$ language for $p>0.5$ because the signal space is restricted, in a perfectly expressive language all signals are used. With this in mind simulations, were performed using a Semi-Supervised ILM with a $9\times 11\times 12$ agent and with a $10\times 15\times 20$. In fact, across all these examples, the behaviour of the model was largely unchanged. 

\begin{figure*}
    \centering
\begin{tabular}{p{0.17\textwidth} p{0.17\textwidth} p{0.17\textwidth} p{0.17\textwidth} p{0.17\textwidth}}
        \multicolumn{5}{c}{$p=0.5$: $\AA$ and $\BB$ have equal influence}\\[10pt]
$x$&$c$&$s$&$a$&$b$\\
\includegraphics[width=0.17\textwidth]{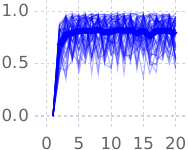}&
\includegraphics[width=0.17\textwidth]{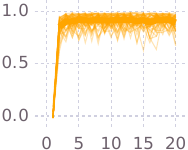}&
\includegraphics[width=0.17\textwidth]{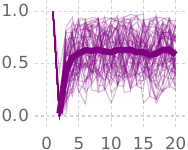}&
\includegraphics[width=0.17\textwidth]{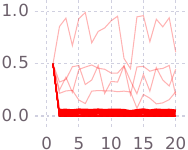}&
\includegraphics[width=0.17\textwidth]{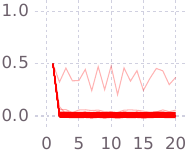}\\[15pt]
\multicolumn{5}{c}{$p=0.75$: $\AA$ has much more influence than $\BB$}\\[10pt]
$x$&$c$&$s$&$a$&$b$\\
\includegraphics[width=0.17\textwidth]{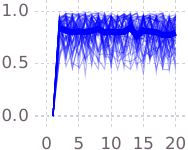}&
\includegraphics[width=0.17\textwidth]{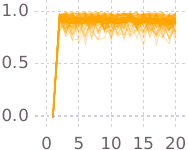}&
\includegraphics[width=0.17\textwidth]{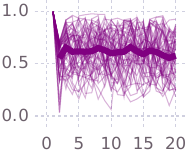}&
\includegraphics[width=0.17\textwidth]{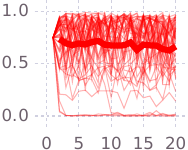}&
\includegraphics[width=0.17\textwidth]{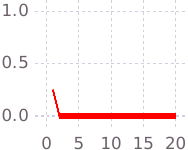}\\
\qquad \quad generations&\qquad \quad generations&\qquad \quad generations&\qquad \quad generations&\qquad \quad generations
    \end{tabular}
    
    \caption{\textbf{Mixing larger languages}. Here the language is evolved for 20 generations and the results of 50 simulations are plotted for the usual five different language attributes: $x$, $c$, $s$, $a$ and $b$. The two rows give two different values of $p$. In each case the encoder and decoder maps have a 20-30-20 layout. 
    }
    \label{fig:large}
\end{figure*}

A much larger language is considered in Fig.~\ref{fig:large}; here a $20\times 30\times 20$ agent is used so the signal and meaning space both have bit length 20. The ILM in this case produces languages which are both expressive and compositional with values near one, as before; however the value of stability fluctuates a lot. Stability is a measure of how many meanings map to the same signal from one generation to the next, in other words it operates at the level of meanings and signals, not facts and morphemes and so it is harsh test of a compositional language: if two bits are swapped, this would reduce stability by a half. In fact, the behaviour seems very similar to the 10-10-10 example, the language quickly becomes expressive and compositional, for the $p=0.5$ case it generally does not resemble either the $\AA$ or $\BB$ language, while for $p=0.75$ it largely resembles $\AA$.

\section{Discussion}

Language contact is complex, both in its internal dynamics and in the broader contexts in which language mixing occurs. The political and culture setting for language contact can be extremely consequential for the populations involved. Language loss can be traumatic and destructive and is often accompanied by other challenging forms of loss; conversely, culture exchange can be enriching and exciting and a useful mechanism under which languages become capable of more efficiently expressing new ideas. None of this complexity is included in the work presented here. The agent modeling also ignores bilingualism and code-switching.  However, what has been demonstrated is that a mechanism which can explain the development of expressivity and compositionality in languages can also explain the stability of languages during language contact, even without invoking the stabilizing influence of a population of speakers not involved in contact.

Recently, population-based versions of the ILM have been considered \citep{BraceBullock2015,BraceBullock2016,SainsHoughtonBullock2023} and a population approach is implicit in the simulations described here in that the initial tutor, which provides examples to the first pupil agent, draws those examples from two different languages. However, this is an extremely restricted version of the richer population dynamics that studying the problem of language contact in more detail would require. In fact, because the Semi-Supervised ILM does not have a sharp distinction between its learning and teaching phases, it lends itself to population models and these will form the basis of further work.

These modeling-based approaches also have a counterpart in participant studies using artificial mini-languages, where similar issues of expressivity, compositionality and stability are also considered, see \cite{FedzechkinaNewportJaeger2016,SmithEtAl2017,CulbertsonSchuler2019} for reviews. A full exploration of language contact should incorporate historical, often very qualitatively evidence, along with simulations and matched data from participant studies.

\section{Conclusion}

In this paper we employed a new class of ILM to explore the outcome of contact between two mature languages that are already expressive, compositional and stable. The semi-supervised nature of this new iterated learning paradigm allows the model to dispense with the computationally expensive and unrealistic obverter mechanism that has tended to limit the size and complexity of languages explored in previous ILM studies. Here, for the first time we are able to model binary languages with up to $2^{20}$ possible meanings and signals. 

The results here demonstrate that when the two pre-contact languages contribute equally to the language exposure of a na\"ive pupil, the resultant language often does not resemble either of the pre-contact languages, and that this is particularly true for larger meaning-signal spaces where recovery of either of the original languages is very rare. However, when one language has some degree of advantage over the other in terms of its contribution to the learning experiences of the na\"ive pupil, this language or something closely resembling it is often able to re-emerge after a period of language destabilization during language contact. This demonstrates that the same dynamics that drives languages to compostionality and expressivity can also stabilize them during language contact.

\section{Acknowledgements}
SB is supported by UKRI grant EP/Y028392/1: AI for Collective Intelligence (AI4CI). Simulations were carried out using the computational facilities of the Advanced Computing Research Centre at the University of Bristol, \texttt{www.bristol.ac.uk/acrc/}. We are grateful to Dr Stewart whose philanthropy supported some of the compute resource used in this project.

\end{document}